\documentclass[12pt]{article}

\usepackage{sbc-template}

\usepackage{graphicx,url}

\usepackage[brazil,english]{babel}
\usepackage[utf8]{inputenc}

\graphicspath{{./images/}{./images/}}
\usepackage{amssymb}  
\usepackage[ruled,vlined,linesnumbered]{algorithm2e}
\usepackage{amsmath} 
\usepackage{float}
\usepackage{subfig}
\usepackage{ctable}

\title{Simple Interactive Image Segmentation using Label Propagation through kNN graphs}

\author{Fabricio A. Breve\inst{1}}
\address{Instituto de Geociências e Ciências Exatas \\ Universidade Estadual Paulista ``Júlio de Mesquita Filho''
  (UNESP)\\
  Avenida 24A, 1515 -- 13.506-900 -- Rio Claro -- SP -- Brazil
  \email{fabricio@rc.unesp.br}
}

\begin{document}

\maketitle

\begin{abstract}
Many interactive image segmentation techniques are based on semi-supervised learning. The user may label some pixels from each object and the SSL algorithm will propagate the labels from the labeled to the unlabeled pixels, finding object boundaries. This paper proposes a new SSL graph-based interactive image segmentation approach, using undirected and unweighted kNN graphs, from which the unlabeled nodes receive contributions from other nodes (either labeled or unlabeled). It is simpler than many other techniques, but it still achieves significant classification accuracy in the image segmentation task. Computer simulations are performed using some real-world images, extracted from the Microsoft GrabCut dataset. The segmentation results show the effectiveness of the proposed approach.
\end{abstract}


\section{Introduction}

Image segmentation is considered one of the most difficult tasks in image processing \cite{Gonzalez2008}. It is the process of dividing an image into parts, identifying objects or other relevant information \cite{Shapiro2001}. Fully automatic segmentation is still very difficult to accomplish and the existing techniques are usually domain-dependent. Therefore, interactive image segmentation, in which the segmentation process is partially supervised, has experienced increasing interest in the last decades \cite{Boykov2001,Grady2006,Protiere2007,Blake2004,Ducournau2014,Ding2010,Rother2004,Paiva2010,Li2010,Artan2010,Artan2011,Xu2008,Breve2015IJCNN,Breve2015ICCSA}.

Semi-supervised learning (SSL) is an important field in machine learning, usually applied when unlabeled data is abundant but the process of labeling is expensive, time consuming and/or requiring intensive work of human specialists \cite{Zhu2005,Chapelle2006}. This characteristics makes SSL an interesting approach to perform interactive image segmentation, which may be seen as a pixel classification process. In this scenario, there are often many unlabeled pixels to be classified. An human specialist can easily classify some of them, which are away from the borders, but the process of defining the borders manually is difficult and time consuming.

Many interactive image segmentation techniques are, in fact, based on semi-supervised learning. The user may label some pixels from each object, away from the boundaries where the task is easier. Then, the SSL algorithm will iteratively propagate the labels from the labeled pixels to the unlabeled pixels, finding the boundaries.
%
This paper proposes a different SSL-based interactive image segmentation approach. It is simpler than many other techniques, but it still achieves significant classification accuracy in the image segmentation task. In particular, it was applied to some real-world images, including some images extracted from the Microsoft GrabCut dataset \cite{Rother2004}. The segmentation results show the effectiveness of the proposed approach.

\subsection{Related work}
The approach proposed in this paper may be classified in the category of graph-based semi-supervised learning. Algorithms on this category rely on the idea of building a graph which nodes are data items (both labeled and unlabeled) and the edges represent similarities between them. Label information from the labeled nodes is propagate through the graph to classify all the nodes \cite{Chapelle2006}. Many graph-based methods \cite{Blum2001,Zhu2003,Zhou2004,Belkin2004,Belkin2005,Joachims2003} are similar and share the same regularization framework \cite{Zhu2005}. They usually employ weighted graphs and labels are spread globally, differently from the proposed approach, where the label spreading is limited to neighboring nodes and the graph is undirected and unweighted.

Another graph-based method, known as Label Propagation through Linear Neighborhoods \cite{Wang2008}, also uses a $k$-nearest neighbors graph to propagate labels. However, the edges have weights, which require the resolution of quadratic programming problems to be calculated, prior to the iterative label propagation process. On the other hand, the proposed approach uses only unweighted edges.

\subsection{Technique overview}
In the proposed method, an unweighted and undirected graph is generated by connecting each node (data item) to its $k$-nearest neighbors. Then, in a iterative process, unlabeled nodes will receive contributions from all its neighbors (either labeled or unlabeled) to define their own label. The algorithm usually converges quickly, and each unlabeled node is labeled after the class from which it received most contributions. Differently from many other graph-based methods, no calculation of edge weights or Laplacian matrix are required.

\section{The Proposed Model}

In this section, the proposed technique will be detailed. Given a bidimensional digital image, the set of pixels are reorganized as $\mathfrak{X} = \{x_1,x_2,\dots,x_L,x_{L+1},\dots,x_N\}$, such that $\mathfrak{X}_{L} = \{x_i\}_{i=1}^{L}$ is the labeled pixel subset and $\mathfrak{X}_{U} = \{x_i\}_{i=L+1}^{N}$ is the unlabeled pixels subset. $\mathfrak{L} = \{1,\dots,C\}$ is the set containing the labels. $y: \mathfrak{X} \rightarrow \mathfrak{L}$ is the function associating each $x_i \in \mathbf{\chi}$ to its label $y(x_i)$ as the algorithm output. The algorithm will estimate $y(x_i)$ for each unlabeled pixel $x_i \in \mathfrak{X}_{U}$.

\subsection{$k$-NN Graph Generation}
\label{sec:GraphGeneration}

A large amount of features may be extracted from each pixel $x_i$ to build the graph. In this paper, $23$ features are used. They are shown on Table~\ref{tab:Features}. These are the same features used in \cite{Breve2015WVC}.

\begin{table}
\centering
\caption{List of features extracted from each image to be segmented}
\begin{tabular}{rl}
  \toprule
  \# & Feature Description \\
  \midrule
  1 & Pixel row location \\
  2 & Pixel column location \\
  3 & Red (R) component of the pixel \\
  4 & Green (G) component of the pixel \\
  5 & Blue (B) component of the pixel \\
  6 & Hue (H) component of the pixel \\
  7 & Saturation (S) component of the pixel \\
  8 & Value (V) component of the pixel \\
  9 & ExR component of the pixel \\
  10 & ExG component of the pixel \\
  11 & ExB component of the pixel \\
  12 & Average of R on the pixel and its neighbors (MR) \\
  13 & Average of G on the pixel and its neighbors (MG) \\
  14 & Average of B on the pixel and its neighbors (MB) \\
  15 & Standard deviation of R on the pixel and its neighbors (SDR) \\
  16 & Standard deviation of G on the pixel and its neighbors (SDG) \\
  17 & Standard deviation of B on the pixel and its neighbors (SDB) \\
  18 & Average of H on the pixel and its neighbors (MH) \\
  19 & Average of S on the pixel and its neighbors (MS) \\
  20 & Average of V on the pixel and its neighbors (MV) \\
  21 & Standard deviation of H on the pixel and its neighbors (SDH) \\
  22 & Standard deviation of S on the pixel and its neighbors (SDS) \\
  23 & Standard deviation of V on the pixel and its neighbors (SDV) \\
  \bottomrule
\end{tabular}
\label{tab:Features}
\end{table}

For measures $12$ to $23$, the pixel neighbors are the $8$-connected neighborhood, except on the borders where no wraparound is applied. All components are normalized to have mean $0$ and standard deviation $1$. They are also scaled by a vector of weights $\lambda$ in order to emphasize/deemphasize each feature during the graph generation. ExR, ExG, and ExB components are obtained from the RGB components using the method described in \cite{Lichman2013}. The HSV components are obtained from the RGB components using the method described in \cite{Smith1978}.

The undirected and unweighted graph is defined as $\mathbf{G} = (\mathbf{V},\mathbf{E})$, where $\mathbf{V} = \{v_1,v_2,\dots,v_N\}$ is the set of nodes, and $\mathbf{E}$ is the set of edges $(v_i, v_j)$. Each node $v_i$ corresponds to a pixel $x_i$. Two nodes $v_i$ and $v_j$ are connected if $v_j$ is among the $k$-nearest neighbors of $v_i$, or vice-versa, considering the Euclidean distance between $x_i$ and $x_j$ features. Otherwise, $v_i$ and $v_j$ are disconnected.

\subsection{Label Propagation}

For each node $v_i$, a domination vector $\mathbf{v_i^\omega(t)} = \{v_i^{\omega_1}(t), v_i^{\omega_2}(t), \dots, v_i^{\omega_C}(t) \}$ is created. Each element $v_i^{\omega_c}(t) \in [0, 1]$ corresponds to the domination level from the class $c$ over the node $v_i$. The sum of the domination vector in each node is always constant, $\sum_{c=1}^{C} v_i^{\omega_c} = 1$.

The domination levels are constant in nodes corresponding to labeled pixels, with full domination by the corresponding class. On the other hand, domination levels are variable in nodes corresponding to unlabeled pixels and they are initially set equally among classes. Therefore, for each node $v_i$, the domination vector $\mathbf{v_i^\omega}$ is set as follows:
\begin{equation}\label{eq:NodesInit}
    v_i^{\omega_c}(0) = \left\{
    \begin{array}{ccl}
        1 & & \mbox{if $x_i$ is labeled and $y(x_i) = c$} \\
        0 & & \mbox{if $x_i$ is labeled and $y(x_i) \neq c$} \\
        \frac{1}{C} & & \mbox{if $x_i$ is unlabeled}
    \end{array}\right..
\end{equation}

In the iterative phase, at each iteration each unlabeled node will get contributions from all its neighbors to calculate its new domination levels. Thus, for each unlabeled node $v_i$, the domination levels are updated as follows:
\begin{equation}\label{eq:NodesUpdate}
    \mathbf{v_i^\omega(t+1)} = \frac{1}{K} \sum_{j \in N(v_i)} \mathbf{v_j^\omega(t)},
\end{equation}
where $K = |N(v_i)|$ is the size of $N(v_i)$, and $N(v_i)$ is the set of the $v_i$ neighbors. In this way, the new dominance vector $\mathbf{v_i^\omega}$ is the arithmetic mean of all its neighbors dominance vectors, no matter if they are labeled or unlabeled.

The average maximum domination levels is defined as follows:
\begin{equation}\label{eq:StopCrit}
    \langle v_i^{\omega_{m}} \rangle, m=\arg\max_c v_i^{\omega_c},
\end{equation}
considering all $v_i$ representing unlabeled nodes. $\langle v_i^{\omega_{m}} \rangle$ is checked every $10$ iterations and the algorithm stops when its increase is below $0.001$ between checkpoints.

At the end of the iterative process, each unlabeled pixel is assigned to the class that has the highest domination level on it:
\begin{equation}\label{eq:LabelUnlabeled}
    y(x_i) = \arg\max_c v_i^{\omega_c}
\end{equation}

\subsection{The Algorithm}
\label{sec:Algorithm}

Overall, the proposed algorithm can be outlined as follows:

\begin{algorithm}[h] \small
  Build the $k$-NN graph, as described in Subsection \ref{sec:GraphGeneration}\;
  Set nodes' domination levels by using Eq. \eqref{eq:NodesInit}\;
  \Repeat{the stopping criterion is satisfied}
  {
    \For{each unlabeled node}
    {
        Update node domination levels by using Eq. \eqref{eq:NodesUpdate}\;
    }
  }
  Label each unlabeled pixel using Eq. \eqref{eq:LabelUnlabeled};
  \caption{The proposed method algorithm}
  \label{alg:Algorithm}
\end{algorithm}

\section{Implementation}
In order to reduce the computational resources required by the proposed method, the following implementation strategy is applied.

The iterative step of the algorithm is very fast in comparison with the graph generation step, i.e., the graph generation dominates the execution time. Therefore, the graph is generated using the \emph{k}-d trees method \cite{Friedman1977}, so the algorithm runs in linearithmic time ($O(N \log N)$).

In the iterative step, each iteration runs in $O(uk)$, where $u$ is the amount of unlabeled nodes and $k$ is usually proportional to the amount of neighbors each node has (not equal because the graph is undirected). $u$ is usually a fraction of $N$ in practical problems, and often $k \ll n$. By increasing $k$, one also increases each iteration execution time. On the other hand, the amount of iterations required to converge decreases as the graph becomes more connected and the labels propagate faster, as it was empirically observed in computer simulations.

The iterative steps are synchronous, i.e., the contributions any node receives to produce its domination vector in time $t+1$ refer to the domination levels its neighbors had in time $t$. Therefore, parallelization of this step, corresponding to the inner loop in steps $4$ and $5$ of the Algorithm \ref{alg:Algorithm}, is possible. Nodes can calculate their new domination vectors in parallel without running into race conditions. Synchronization is only required between iterations of the outer loop (steps $3$ to $6$).

\section{Experiments}
\label{sec:experiments}
\begin{figure}
\centering
\subfloat[]{
\includegraphics[height=4cm]{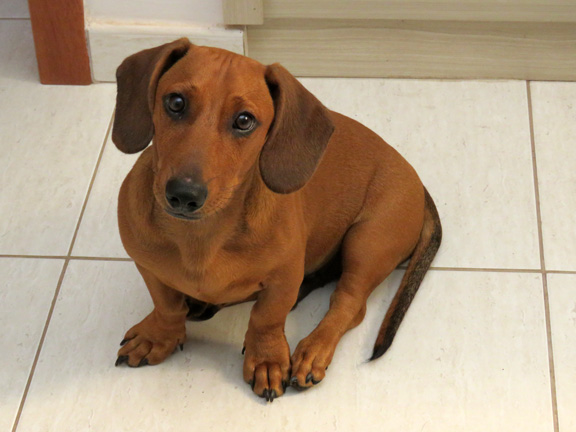}
\label{fig:ralph-org}}
\subfloat[]{
\includegraphics[height=4cm]{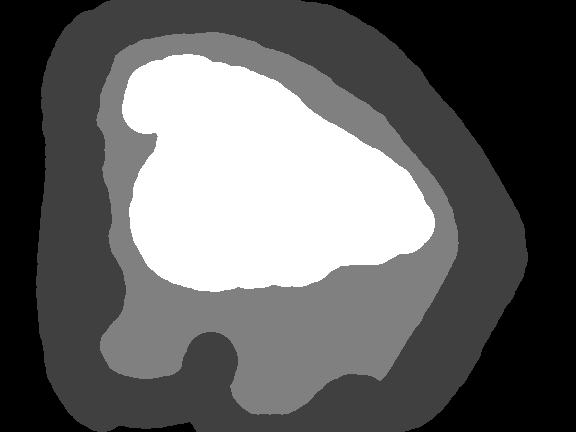}
\label{fig:ralph-slab}
}
\subfloat[]{
\includegraphics[height=4cm]{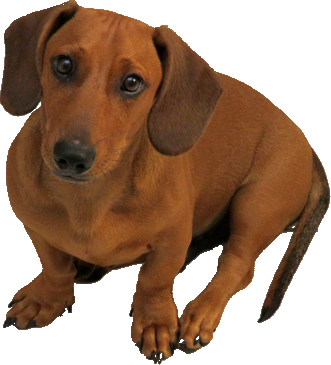}
\label{fig:ralph-res}
}
\caption{(a) Original image, (b) Trimap providing seed regions for Fig.~\ref{fig:ralph-org} segmentation, (c) Close-up foreground segmentation results by the proposed method.}
\label{fig:ralph}
\end{figure}

The proposed technique efficacy is first tested using the real-world image shown on Fig.~\ref{fig:ralph-org}, extracted from \cite{Breve2015IJCNN}, which has $576 \times 432$ pixels. A trimap providing seed regions is presented in Figure \ref{fig:ralph-slab}. Black (0) represents the background, ignored by the algorithm; dark gray (64) is the labeled background; light gray (128) is the unlabeled region, which labels will be estimated by the proposed method; and white (255) is the labeled foreground.

\begin{figure}
\centering
\begin{tabular}{ccc}
\subfloat[]{\includegraphics[width=4.0cm,height=4.0cm,keepaspectratio]{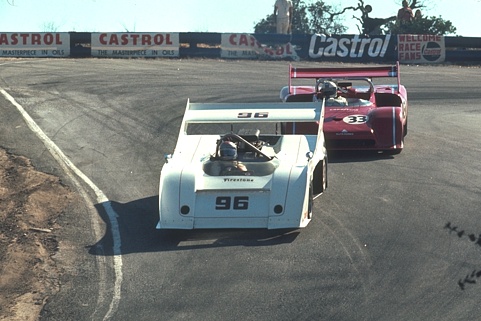}} &
\subfloat[]{\includegraphics[width=4.0cm,height=4.0cm,keepaspectratio]{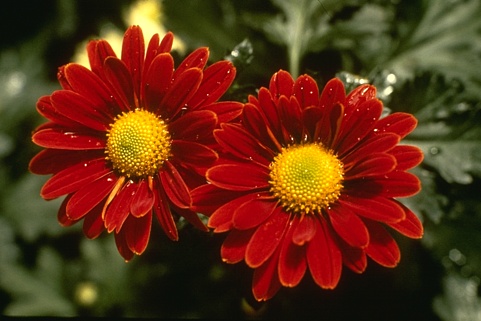}} &
\subfloat[]{\includegraphics[width=4.0cm,height=4.0cm,keepaspectratio]{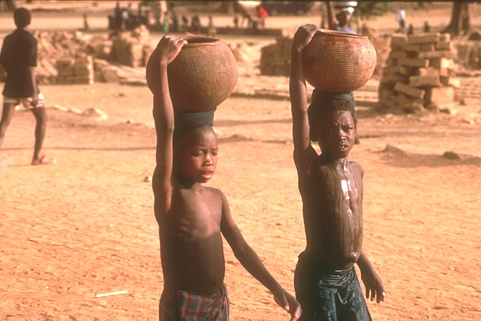}} \\
\subfloat[]{\includegraphics[width=4.0cm,height=4.0cm,keepaspectratio]{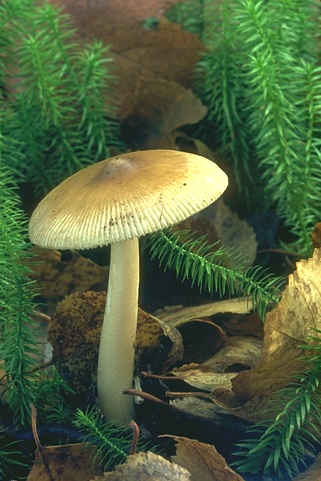}} &
\subfloat[]{\includegraphics[width=4.0cm,height=4.0cm,keepaspectratio]{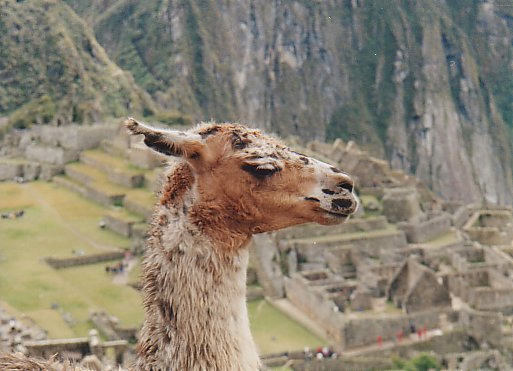}} &
\subfloat[]{\includegraphics[width=4.0cm,height=4.0cm,keepaspectratio]{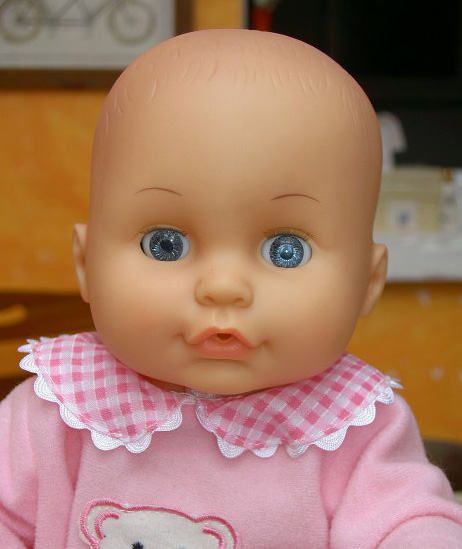}}\\
\subfloat[]{\includegraphics[width=4.0cm,height=4.0cm,keepaspectratio]{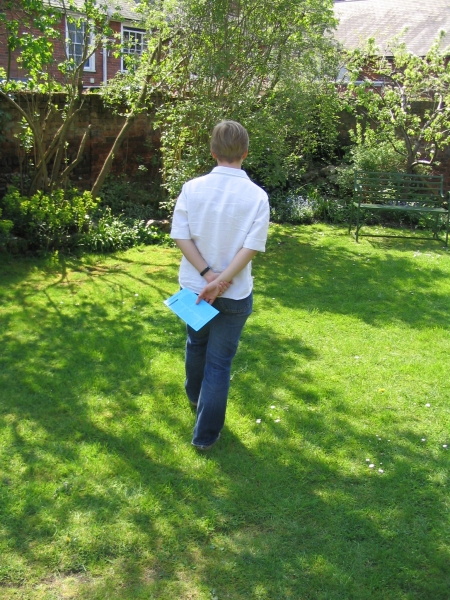}} &
\subfloat[]{\includegraphics[width=4.0cm,height=4.0cm,keepaspectratio]{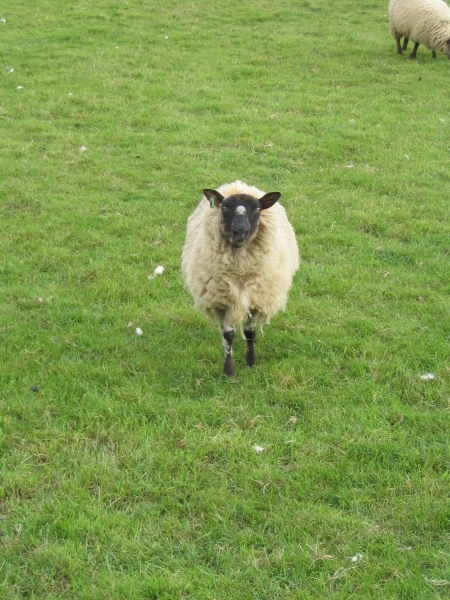}} &
\subfloat[]{\includegraphics[width=4.0cm,height=4.0cm,keepaspectratio]{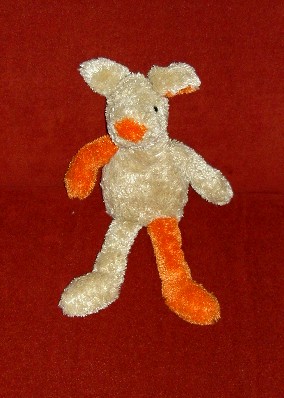}}
\end{tabular}
\caption{Original images from the GrabCut dataset: (a) 21077; (b) 124084; (c) 271008; (d) 208001; (e) llama; (f) doll; (g) person7; (h) sheep; (i) teddy.}
\label{fig:grabcut-images}
\end{figure}

\begin{figure}
\centering
\begin{tabular}{ccc}
\subfloat[]{\includegraphics[width=4.0cm,height=4.0cm,keepaspectratio]{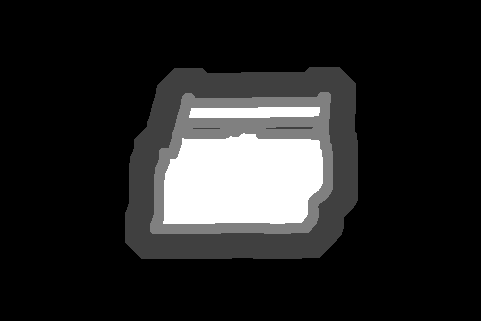}} &
\subfloat[]{\includegraphics[width=4.0cm,height=4.0cm,keepaspectratio]{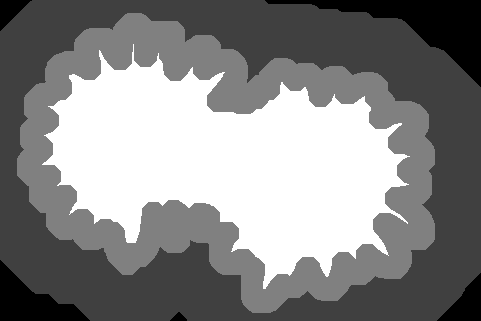}} &
\subfloat[]{\includegraphics[width=4.0cm,height=4.0cm,keepaspectratio]{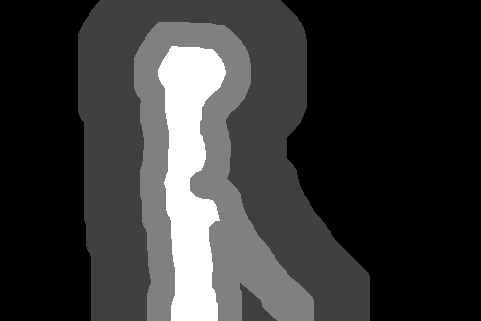}} \\
\subfloat[]{\includegraphics[width=4.0cm,height=4.0cm,keepaspectratio]{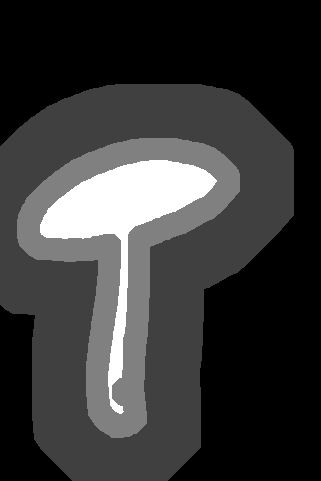}} &
\subfloat[]{\includegraphics[width=4.0cm,height=4.0cm,keepaspectratio]{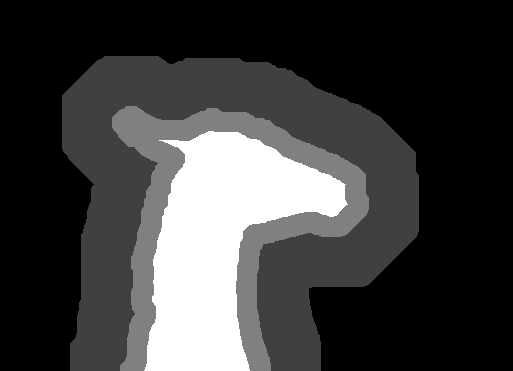}} &
\subfloat[]{\includegraphics[width=4.0cm,height=4.0cm,keepaspectratio]{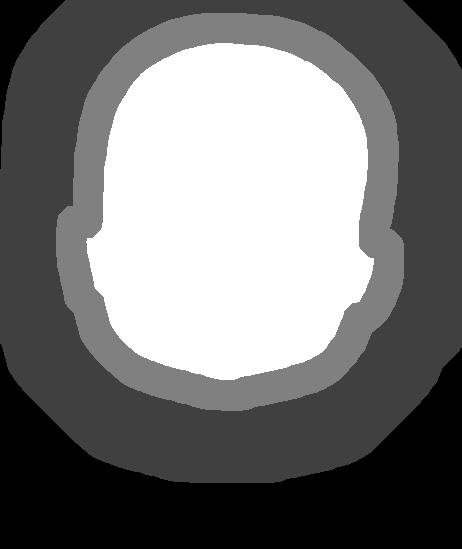}}\\
\subfloat[]{\includegraphics[width=4.0cm,height=4.0cm,keepaspectratio]{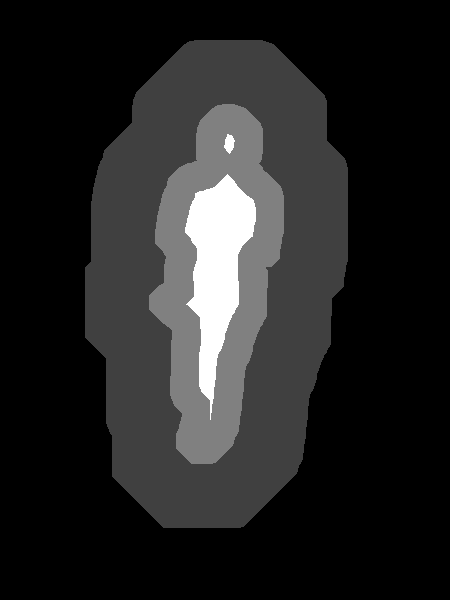}} &
\subfloat[]{\includegraphics[width=4.0cm,height=4.0cm,keepaspectratio]{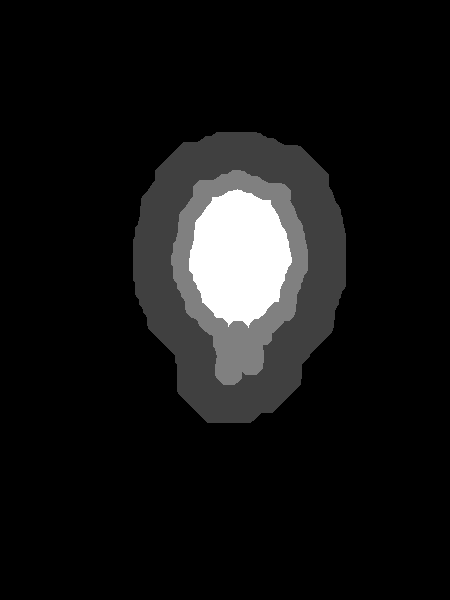}} &
\subfloat[]{\includegraphics[width=4.0cm,height=4.0cm,keepaspectratio]{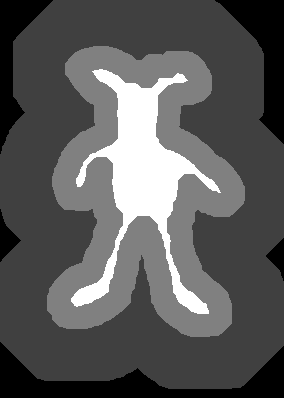}}
\end{tabular}
\caption{The trimaps providing seed regions from the GrabCut dataset: (a) 21077; (b) 124084; (c) 271008; (d) 208001; (e) llama; (f) doll; (g) person7; (h) sheep; (i) teddy.}
\label{fig:grabcut-lasso}
\end{figure}

\begin{figure}
\centering
\begin{tabular}{ccc}
\subfloat[]{\includegraphics[width=4.0cm,height=4.0cm,keepaspectratio]{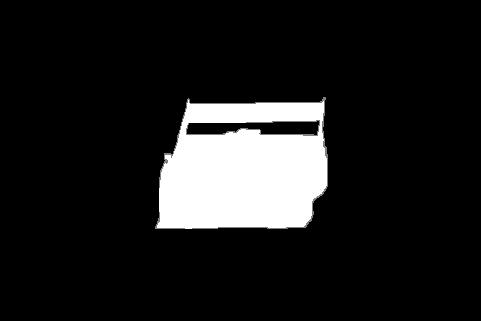}} &
\subfloat[]{\includegraphics[width=4.0cm,height=4.0cm,keepaspectratio]{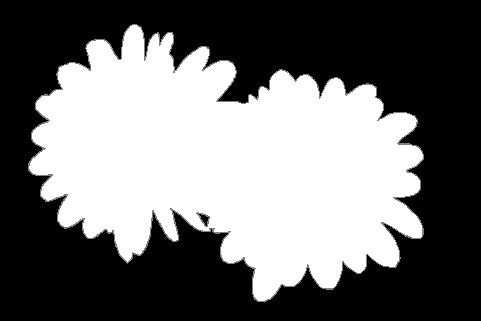}} &
\subfloat[]{\includegraphics[width=4.0cm,height=4.0cm,keepaspectratio]{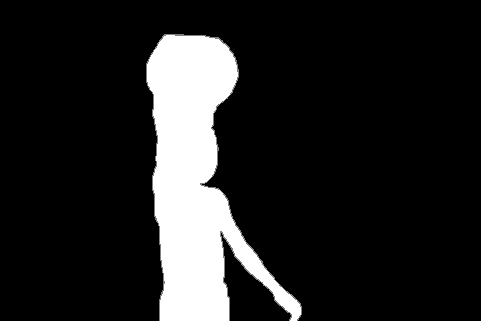}} \\
\subfloat[]{\includegraphics[width=4.0cm,height=4.0cm,keepaspectratio]{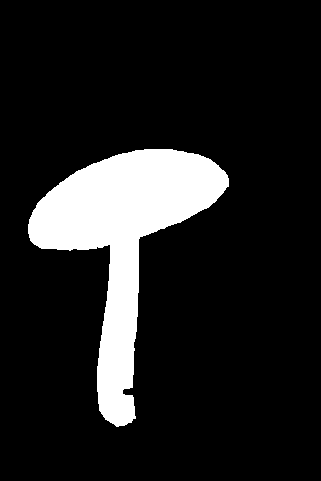}} &
\subfloat[]{\includegraphics[width=4.0cm,height=4.0cm,keepaspectratio]{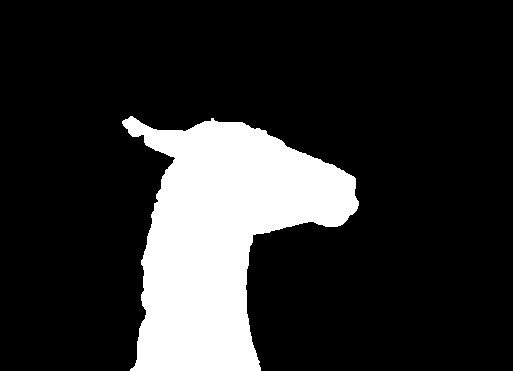}} &
\subfloat[]{\includegraphics[width=4.0cm,height=4.0cm,keepaspectratio]{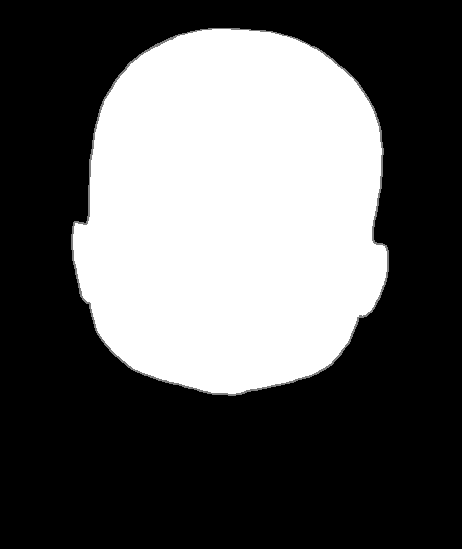}}\\
\subfloat[]{\includegraphics[width=4.0cm,height=4.0cm,keepaspectratio]{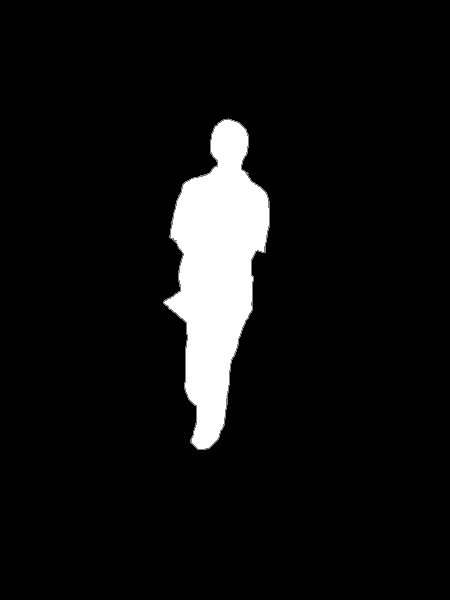}} &
\subfloat[]{\includegraphics[width=4.0cm,height=4.0cm,keepaspectratio]{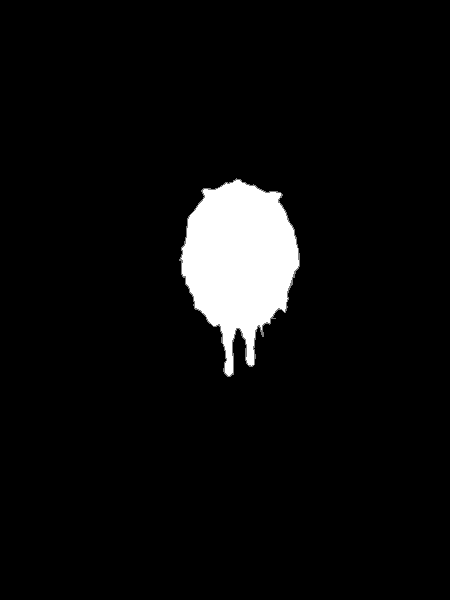}} &
\subfloat[]{\includegraphics[width=4.0cm,height=4.0cm,keepaspectratio]{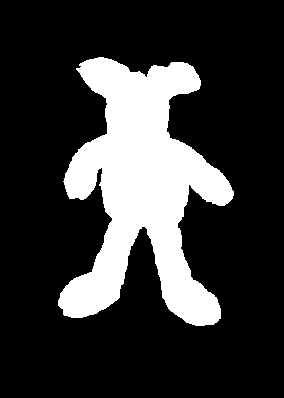}}
\end{tabular}
\caption{Close-up foreground segmentation results by the proposed method: (a) 21077; (b) 124084; (c) 271008; (d) 208001; (e) llama; (f) doll; (g) person7; (h) sheep; (i) teddy.}
\label{fig:grabcut-gt}
\end{figure}

The proposed technique efficacy is then verified using a series of computational experiments using nine image selected from the Microsoft GrabCut database \cite{Rother2004}  \footnote{Available at \url{http://web.archive.org/web/20161203110733/research.microsoft.com/en-us/um/cambridge/projects/visionimagevideoediting/segmentation/grabcut.htm}}. The selected images are shown on Fig.~\ref{fig:grabcut-images}. The corresponding \emph{trimaps} providing seed regions are shown on Fig.~\ref{fig:grabcut-lasso}. Finally, the \emph{ground truth} images are shown on Fig.~\ref{fig:grabcut-gt}.

For each image, $k$ and the vector of weights $\lambda$ were optimized using the genetic algorithm available in Global Optimization Toolbox of MATLAB, with its default parameters.

%
%
%
%

\section{Results and Discussion}

First, the proposed method was applied to the image shown on Fig.~\ref{fig:ralph-org}. The best segmentation result is shown on Fig.~\ref{fig:ralph-res}. By comparing this output with the segmentation result achieved in \cite{Breve2015IJCNN} for the same image, one can notice that the proposed method achieved slightly better results, by eliminating some misclassified pixels and better defining the borders.

\begin{figure}
\centering
\begin{tabular}{ccc}
\subfloat[5.13\%]{\includegraphics[width=4.0cm,height=4.0cm,keepaspectratio]{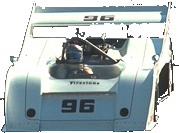}} &
\subfloat[0.57\%]{\includegraphics[width=4.0cm,height=4.0cm,keepaspectratio]{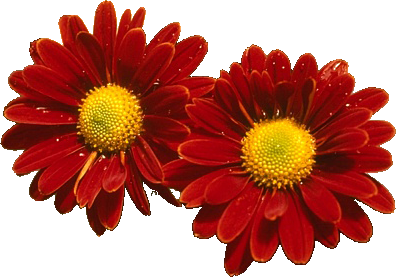}} &
\subfloat[3.09\%]{\includegraphics[width=4.0cm,height=4.0cm,keepaspectratio]{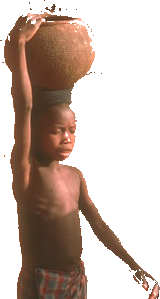}} \\
\subfloat[3.88\%]{\includegraphics[width=4.0cm,height=4.0cm,keepaspectratio]{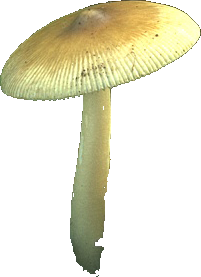}} &
\subfloat[6.83\%]{\includegraphics[width=4.0cm,height=4.0cm,keepaspectratio]{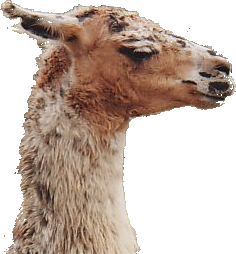}} &
\subfloat[0.64\%]{\includegraphics[width=4.0cm,height=4.0cm,keepaspectratio]{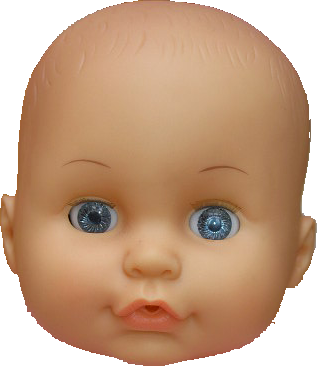}}\\
\subfloat[1.09\%]{\includegraphics[width=4.0cm,height=4.0cm,keepaspectratio]{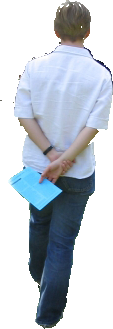}} &
\subfloat[1.41\%]{\includegraphics[width=4.0cm,height=4.0cm,keepaspectratio]{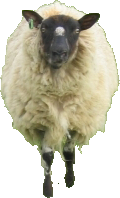}} &
\subfloat[1.63\%]{\includegraphics[width=4.0cm,height=4.0cm,keepaspectratio]{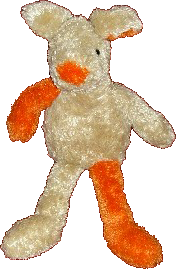}}
\end{tabular}
\caption{The ground-truth images from the GrabCut dataset: (a) 21077; (b) 124084; (c) 271008; (d) 208001; (e) llama; (f) doll; (g) person7; (h) sheep; (i) teddy. Error rates are indicated below each image.}
\label{fig:grabcut-res}
\end{figure}

Then, the proposed method was applied to the nine images shown on Fig.~\ref{fig:grabcut-images}, as described on Section \ref{sec:experiments}. The best segmentation results achieved with the proposed method are shown on Fig.~\ref{fig:grabcut-res}. Error rates are computed as the fraction between the amount of incorrectly classified pixels and the total amount of unlabeled pixels (light gray on the \emph{trimaps} images shown on Fig.~\ref{fig:grabcut-lasso}). Notice that \emph{ground truth} images (Fig.~\ref{fig:grabcut-gt}) have a thin contour of gray pixels, which corresponds to uncertainty, i.e., pixels that received different labels by the different persons who did the manual classification. These pixels are not used in the classification error calculation.

Segmentation error rates are also summarized on Table~\ref{tab:error}. Some results from other methods \cite{Ducournau2014,Ding2008,Breve2015IJCNN} are also included for reference. By analyzing them, one can notice that the proposed method has comparable results. The results from the other methods were extracted from the respective references.

\begin{table}[H]
  \centering
  \caption{Segmentation error rates achieved by Learning on Hypergraphs model (ISLH) \cite{Ding2008}, Directed Image Neighborhood Hypergraph model (DINH) \cite{Ducournau2014}, Particle Competition and Cooperation (PCC) \cite{Breve2015IJCNN} and the proposed model (LPKNN).}
  \includegraphics[width=8cm]{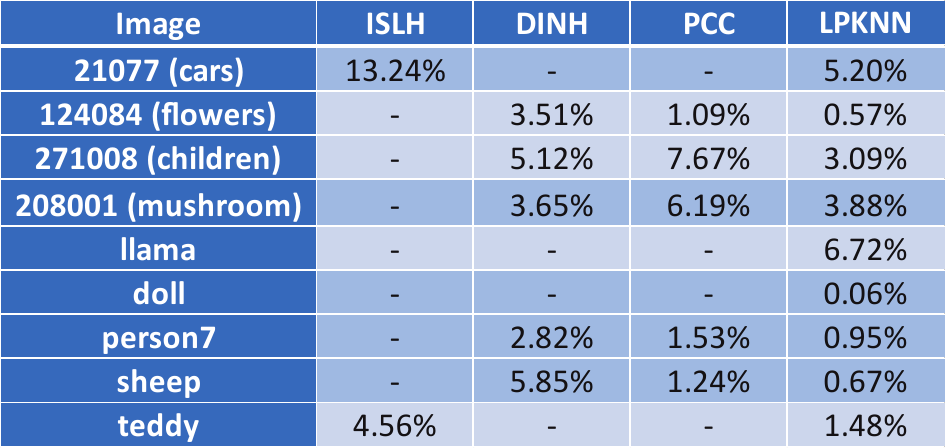}
  \label{tab:error}
\end{table}

It is also important to notice that the proposed method is deterministic. Given the same parameters, it will always output the same segmentation result on different executions. Other methods, like Particle Competition and Cooperation \cite{Breve2015IJCNN}, are stochastic. Therefore, they may output different segmentation results on each execution.

The optimized parameters $k$ and features weights ($\lambda$) are shown on Table~\ref{tab:parameters}. Considering the $10$ images evaluated in this paper, pixel location features (Row and Col) are the most important features, followed by the ExB component, intensity (V), and the mean of green (MG). The least important features were hue (H), saturation (S) and all those related to standard deviation. However, no single feature received a high weight in all images. The optimal weights and $k$ seem to be highly dependent on image characteristics.

\begin{table*}
  \centering
  \caption{Parameter $k$ and feature weights $\lambda$ optimized by the proposed method for each segmented image.}
  \includegraphics[width=15cm]{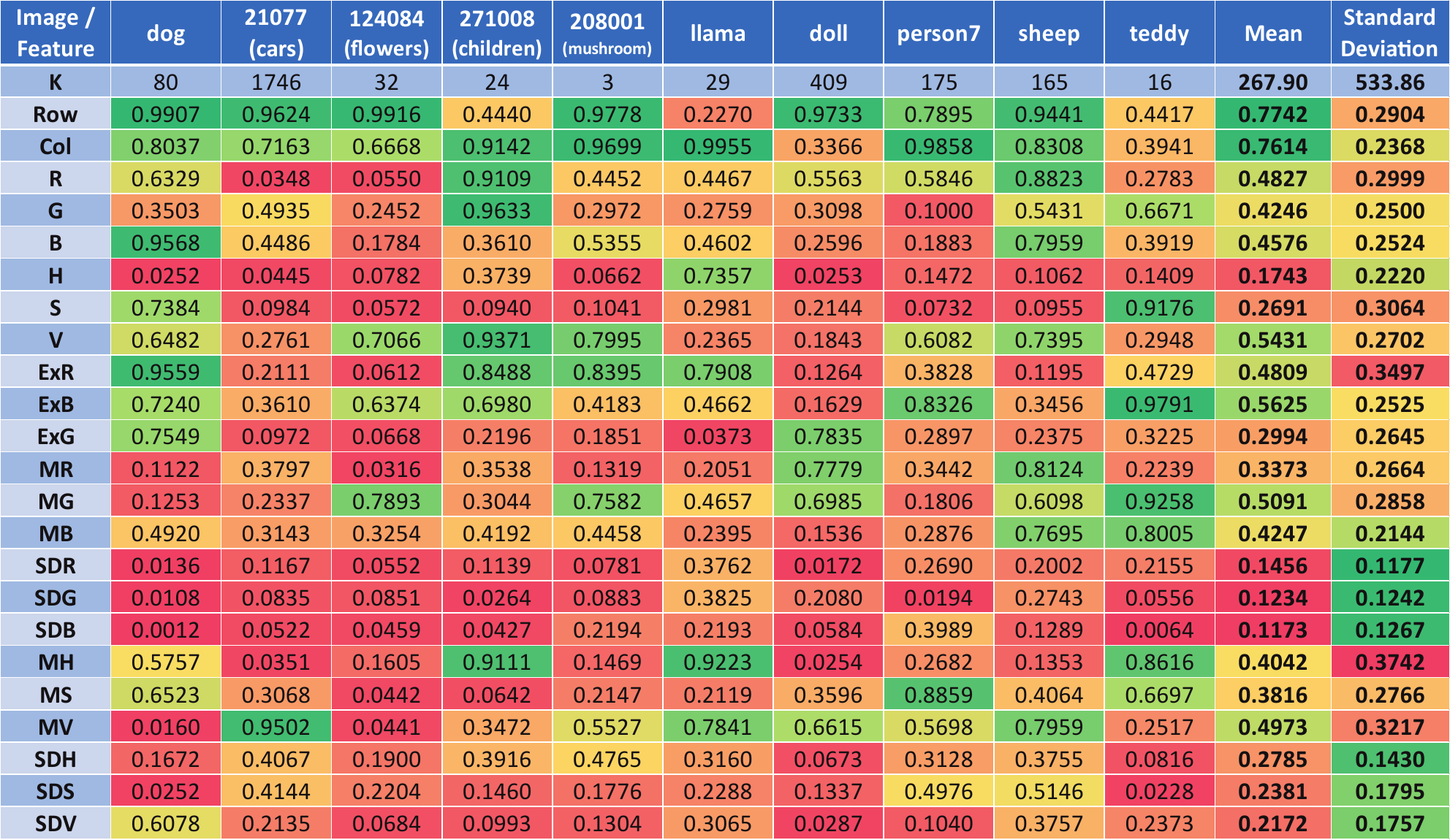}
  \label{tab:parameters}
\end{table*}

\section{Conclusion}
In this paper, a new SSL graph-based approach is proposed to perform interactive image segmentation. It employs undirected and unweighted $k$NN graphs to propagate labels from nodes representing labeled pixels to nodes representing unlabeled pixels. Computer simulations with some real-world images show that the proposed approach is effective, achieving segmentation accuracy similar to those achieved by some state-of-the-art methods.

As future work, the method will be applied on more images and more features may be extracted. Methods to automatically define the parameters $k$ and $\lambda$ may also be explored. Graph generation may also be improved to provide further increase in segmentation accuracy.

Moreover, the proposed method works for multiple labels simultaneously at no extra cost, which is an interesting property not often exhibited by other interactive image segmentation methods. This feature will also be explored in future works.

\section*{Acknowledgment}

The author would like to thank the S\~{a}o Paulo Research Foundation - FAPESP (grant \#2016/05669-4) and the National Counsel of Technological and Scientific Development - CNPq (grant \#475717/2013-9) for the financial support.

\bibliographystyle{sbc}
\bibliography{lpknn}

\end{document}